\pdfoutput=1
\documentclass[11pt]{article}

\usepackage[final]{acl}

\usepackage{times}
\usepackage{latexsym}

\usepackage[T1]{fontenc}
\usepackage[utf8]{inputenc}

\usepackage{microtype}

\usepackage{inconsolata}

\usepackage{graphicx}

\usepackage{refstyle}
\usepackage{amsmath}
\usepackage{cleveref}

\usepackage{xspace}
\usepackage{booktabs}
\usepackage{graphicx}
\usepackage{adjustbox}
\usepackage{array}
\usepackage{multirow}
\usepackage{cleveref}
\usepackage[normalem]{ulem}

\usepackage{amssymb}
\usepackage{booktabs}
\usepackage{soul}% for underlines
\usepackage{pifont}
\usepackage{bbm}

\usepackage{graphicx}
\usepackage{caption}
\usepackage{subcaption}
\newcommand{\reducedstrut}{\vrule width 0pt height .9\ht\strutbox depth .9\dp\strutbox\relax}
\newcommand{\purple}[1]{%
  \begingroup
  \setlength{\fboxsep}{0pt}%  
  \colorbox{violet!20}{\reducedstrut#1\/}%
  \endgroup
}
\newcommand{\pink}[1]{%
  \begingroup
  \setlength{\fboxsep}{0pt}%  
  \colorbox{red!20}{\reducedstrut#1\/}%
  \endgroup
}
\newcommand{\skyblue}[1]{%
  \begingroup
  \setlength{\fboxsep}{0pt}%  
  \colorbox{cyan!20}{\reducedstrut#1\/}%
  \endgroup
}
\usepackage{xcolor,pifont}
\newcommand{\greencheck}{{\color{green}\ding{51}}}
\newcommand{\redcross}{{\color{red}\ding{55}}}

\usepackage{arydshln}

\usepackage{tcolorbox}
\usepackage{lipsum} % For dummy text

\crefformat{section}{\S#2#1#3}
\crefformat{subsection}{\S#2#1#3}
\crefformat{subsubsection}{\S#2#1#3}
\crefrangeformat{section}{\S#3#1#4 to~\S#5#2#6}
\crefmultiformat{section}{\S#2#1#3}{ and~\S#2#1#3}{, #2#1#3}{ and~#2#1#3}
\Crefformat{figure}{#2Figure~#1#3}
\Crefmultiformat{figure}{Figs.~#2#1#3}{ and~#2#1#3}{, #2#1#3}{ and~#2#1#3}
\Crefformat{table}{#2Table~#1#3}
\Crefmultiformat{table}{Tabs.~#2#1#3}{ and~#2#1#3}{, #2#1#3}{ and~#2#1#3}
\Crefformat{appendix}{#2Appx.~\S#1#3}
\crefformat{algorithm}{Alg.~#2#1#3}
\Crefformat{equation}{#2Eq.~#1#3}

\newcommand{\stitle}[1]{\vspace{1ex} \noindent{\bf #1.}}

\newcommand{\MODEL}{\mbox{\textsc{FaviComp}}\xspace}

\title{Familiarity-Aware Evidence Compression for \mbox{Retrieval-Augmented Generation}}

\author{
Dongwon Jung\textsuperscript{1} \quad
Qin Liu\textsuperscript{1} \quad 
Tenghao Huang\textsuperscript{2} \quad 
Ben Zhou\textsuperscript{3} \quad 
Muhao Chen\textsuperscript{1} \\
\textsuperscript{1}University of California, Davis,
\textsuperscript{2}University of Southern California,
\textsuperscript{3}Arizona State University \\
\texttt{\{dwojung,qinli,muhchen\}@ucdavis.edu} \quad
\texttt{tenghaoh@usc.edu} \quad
\texttt{benzhou@asu.edu} 
}

\begin{document}
\maketitle

\begin{abstract}
%Retrieval-Augmented Generation (RAG) enhances large language models (LMs) by integrating external knowledge, crucial for knowledge-intensive tasks like open-domain question answering (QA) and multi-hop reasoning. However, traditional RAG methods often struggle with filtering irrelevant information, leading to suboptimal performance. Existing works deal with this challenge by compressing the retrieved evidence set to retraining only the essential information for the LMs to utilize effectively. However, the compressed evidence might be unfamiliar to the LM employed for the downstream task (referred to as target model) due to discrepancies in internal knowledge and prompt preferences between the compressor module and the target model. 
Retrieval-augmented generation (RAG) improves large language models (LMs) by incorporating non-parametric knowledge through evidence retrieved from external sources. However, it often struggles to cope with inconsistent and irrelevant information that can distract the LM from its tasks, especially when multiple evidence pieces are required. While compressing the retrieved evidence with a compression model aims to address this issue, the compressed evidence may still be unfamiliar to the target model used for downstream tasks, potentially failing to utilize the evidence effectively.
We propose \MODEL (\textbf{\textsc{Fa}}miliarity-aware E\textbf{\textsc{vi}}dence \textbf{\textsc{Comp}}ression), a novel inference-time evidence compression technique that makes retrieved evidence more familiar to the target model, while seamlessly integrating parametric knowledge from the model. 
%Specifically, \MODEL proactively composes the compressed evidence in a way to lower the perplexity of the target model by combining decoding probabilities from both the compression model and the target model to generate context that is more familiar to the target model. This approach balances the integration of parametric and non-parametric knowledge, which is especially helpful in complex tasks where the retrieved evidence set may not contain all the necessary information. 
Experimental results show that \MODEL consistently outperforms the most recent evidence compression baselines across multiple open-domain QA datasets, improving accuracy by up to 28.1\% while achieving high compression rates. Additionally, we demonstrate the effective integration of both parametric and non-parametric knowledge during evidence compression. \footnote{Code and data are available at \url{https://github.com/luka-group/FaviComp}}
\end{abstract}

\begin{figure*}[t]
    \centering
    \small\includegraphics[width=0.95\linewidth]{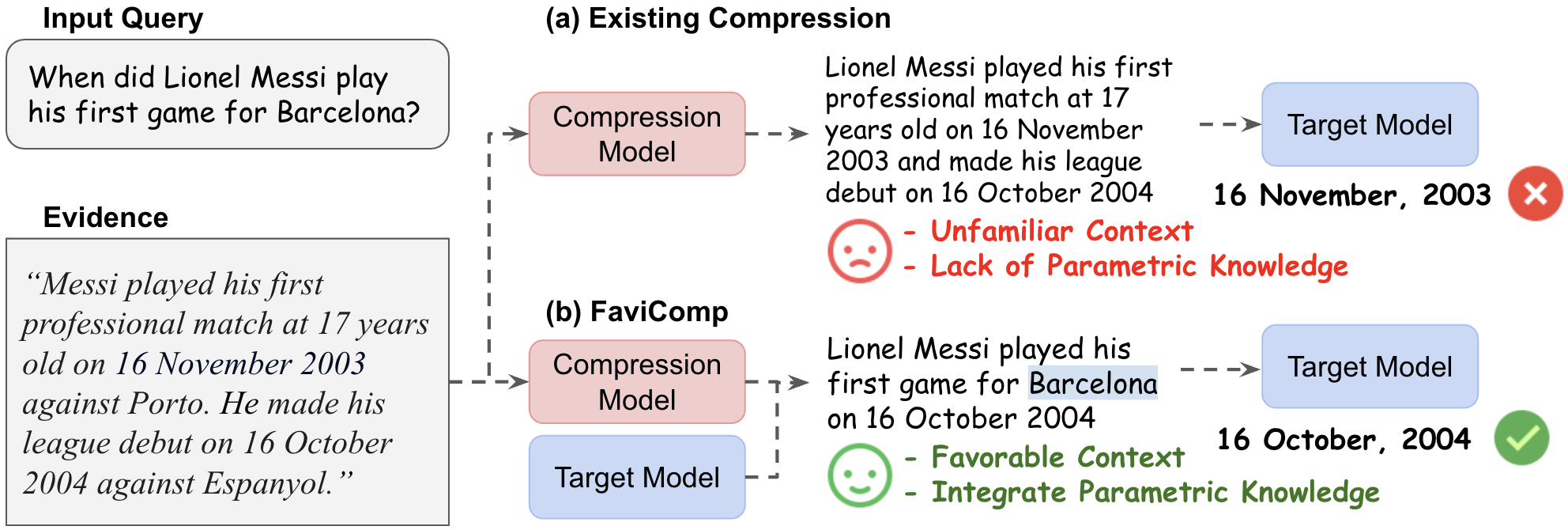}
    \caption{An overview of \MODEL. 
    Instead of relying solely on compressed evidence from the compression model (upper), \MODEL familiarizes the compressed evidence to the target model while integrating parametric knowledge through ensemble decoding, resulting in improved downstream performance (lower).}
    \label{fig:method}
    \vspace{-1em}
\end{figure*}

\section{Introduction}

Retrieval-augmented generation (RAG) has become a common paradigm %to enhance the performance of large language models (LMs) for knowledge-intensive tasks. 
%One of the key paradigms of RAG is the retrieve-then-read approach \citep{lewis2020retrieval}, which involves retrieving textual evidence related to the prompt from the external corpora or knowledge bases and incorporating it as a part of the input to the LM.
%RAG enables LMs 
for large language models (LMs) to leverage external knowledge beyond their inherent knowledge boundaries to perform better in knowledge-intensive tasks such as open-domain question answering (QA) \citep{lewis2020retrieval,izacard-grave-2021-leveraging,guu2020retrieval} and fact-checking \citep{pan2023fact,li2024self}. %\ben{It's unclear what inherent boundaries are. Maybe say something along the line of retrieval offloads some memorization requirement so LMs can work with more accurate information in knowledge-intensive tasks?}
In particular, incorporating multiple evidence pieces is crucial in solving complicated tasks such as multi-hop and complex reasoning \citep{trivedi2023interleaving,jiang2023active,li2024deceptive,lu2023multi}, which require various sources of information to solve the questions.

Nevertheless, RAG often struggles to cope with inconsistent and irrelevant information from the multiple evidence pieces, which can interfere with downstream tasks \citep{shi2023large}. %naturally faces the challenge in which multiple evidence pieces may not collectively form a coherent description of the evidence for the LM. \ben{I don't think the issue is coherency, it should be relevancy? The retrieved docs are technically 'coherent'}
% \qin{description of coherency, e.g. different context, conflict between evidences, ...}
%Often, the retrieved documents provide inconsistent narratives, 
This highlights the need for evidence compression to identify and retain only the essential information for LMs to utilize effectively.
% Particularly, the obtained documents may not provide consistent evidence and may narrate differently. This necessitates \emph{post-retrieval processing} of the documents to identify only essential information for the LMs to utilize. 
Traditionally, evidence compression has focused on reranking documents or sentences by relevance and then incorporating a top-ranked subset \citep{nogueira2020document,zhuang2023rankt5,wang2023learning} %However, this approach risks losing valuable information contained in the discarded documents. Recent studies have explored 
or compressing the documents into a compact form that retains only essential context 
\citep{jiang2023longllmlingua,xu2024recomp,yoon2024compact}.
However, the compressed evidence might be unfamiliar to the LM employed for the downstream task (referred to as the target model), particularly due to discrepancies in the internal knowledge and prompt preferences between the compression model and the target model %that are learned during the pretraining stage 
\citep{gonen2023demystifying,lee2024crafting,li2024famicom, mallen2023not}. 
When LMs encounter unfamiliar contextual information, they often fail in balancing parametric and non-parametric knowledge, either by overly relying on their parametric knowledge \citep{longpre2021entity,wang2023causal,zhou2023context} or by utilizing retrieved evidence without considering its relevance to the input \citep{wu2024clasheval}.
%to use it effectively by overlooking essential details or relying too heavily on their parametric knowledge, leading to sub-optimal downstream performance \citep{longpre2021entity,tan2024blinded}.
% \qin{This passage is claiming two main challenges of RAG. Shall we categorize it as (1) unfamiliarity or potential (2) conflict? If (1), then we can both cite related works and provide experimental results to support this claim by showing how unfamiliarity can harm the results. If (2), then we could probably formalize the challenge into two types of conflicts: intra-evidence conflict and the conflict between retrieved evidences and target LM's parametric knowledge. Besides, if (1) takes efforts to claim, we can also claim the first challenge only (conflicts between evidences), and provide the second one as a bonus instead of challenge.}

To address these challenges, we propose \textbf{\textsc{Fa}}miliarity-aware E\textbf{\textsc{vi}}dence \textbf{\textsc{Comp}}ression (\MODEL), an inference-time evidence compression method that %compresses evidence that is more familiar to the target model while integrating parametric knowledge into the compressed evidence. 
consolidates multiple evidence into an abstractive summary that is more familiar to the target model, while seamlessly integrating parametric knowledge from the model.
Inspired by the prior findings that an LM's familiarity with a prompt is generally reflected by low perplexity  \citep{liu2024monotonic,gonen2023demystifying,wang2023readprompt}, \MODEL proactively %lowers the perplexity of the
composes the compressed evidence in a way to lower the perplexity of the target model.
%We achieve this by restricting the token search space at each decoding step to tokens that yield lower perplexity for the target model \citep{liu2024monotonic}. 
%Specifically, \MODEL leverages the decoding probabilities of two LMs, a \emph{compression model} and the \emph{target model}. The compression model is instructed to summarize the raw evidential documents into a relevant context to the input, while the target model is instructed to generate relevant context without referencing the documents.
Specifically, instead of directly selecting the highest probability token from the compression model at each decoding step, \MODEL selects the token from the ensemble of the token probabilities from both the compression and target models. % and then select the token with the highest probability from this combined set. 
This ensemble decoding therefore constrains the token search space of the compression model to those with lower perplexity for the target model, making the context more familiar to the target model \citep{liu2024monotonic}. 

Furthermore, \MODEL potentially synergizes the retrieved knowledge with the target model's parametric knowledge introduced during ensemble decoding. It can effectively discern when to leverage internal or external knowledge, which is particularly beneficial in the presence of noisy contextual evidence in complex tasks such as multi-document or multi-hop QA \citep{wang2024knowledge}.

%\MODEL presents three key advantages. First, it is an unsupervised refinement method that does not require any training for additional reranker or compressor modules. Second, it compresses the raw documents into a format optimized for the downstream model, thereby enhancing overall performance. Lastly, it effectively integrates internal and external knowledge, which is especially crucial when retrieval is challenging. Given that balancing a model's internal memory and retrieved knowledge is an under-explored research area, our work represents a preliminary investigation in this field.
%\MODEL brings along key advantages of RAG for complex tasks from two perspectives. On the one hand, it is capable of compressing multiple augmented documents to a more favorable form to the target model. This mechanism not only helps the model better comprehend the essential evidence in the retrieval augmentation but also better balances knowledge utility in both the evidential context and the model's parametric memory. On the other hand, it is a training-free and model-agnostic approach that can be easily plugged into any RAG processes % of any auto-regressive LMs.

Our experiments show that \MODEL outperforms %all existing 
most recent evidence compression baselines in five open-domain QA datasets, improving accuracy by up to 28.1\% while maintaining high compression rates. Additionally, we conduct ablation studies by varying the degree of decoding ensemble and analyzing its impact on performance and context perplexity. Moreover, we investigate how \MODEL effectively integrates parametric and non-parametric knowledge during evidence compression.

%Our contributions are as follows: (1) We demonstrate that MonoComp successfully addresses the issues of existing post-retrieval methods in RAG and showcase the synergy of LM's internal memory and externally retrieved context. (2) We conduct extensive experiments to prove the effectiveness of MonoComp on 5 question-answering datasets with two LMs, Llama-3-8B-Instruct and mistral-7B-Instruct-v0.3. (3) We perform additional ablation studies on efficiency 

\section{Method}
%\qin{name of this section is subject to change, based on the title}

We present \MODEL, a inference-time evidence compression method that familiarizes retrieved evidence with the target model while synergizing them with the model's parametric knowledge. We first illustrate the motivation for \MODEL in \Cref{sec:motivation} and provide the preliminaries of evidence compression in RAG \Cref{sec:preliminary}, followed by a detailed definition of 
%introduce the problem settings and then provide
%comprehensive details of 
our proposed framework in \Cref{sec:monocomp}.

\subsection{Motivation and Method Overview}
\label{sec:motivation}

\Cref{fig:method} illustrates the overview of  \MODEL. Existing evidence compression methods employ the compression model to filter out irrelevant information from the retrieved documents. However, since the compression model and the target model are different, the target model might not be familiar to the compressed evidence due to the difference in internal knowledge and prompt preferences between the two models \citep{gonen2023demystifying,lee2024crafting,mallen2023not}. In addition, the compressed evidence cannot be supplemented with the rich parametric knowledge from the target model. In the example, even though the compression model successfully summarizes the essential information, the target model produces an inaccurate answer due to the unfamiliarity with the target model and the lack of integration of the parametric knowledge. On the other hand, \MODEL compresses the given evidence more favorable to the target model by using a novel ensemble decoding technique and leverages its parametric knowledge to supplement the missing evidence (\textit{``Lionel Messi made his league debut in Barcelona"}), effectively combining evidential and parametric knowledge.

% \subsection{Problem Formulation}
% \label{sec:problem}
% Given a set of $k$ retrieved evidence $D=\{d_1,d_2,\ldots, d_k\}$ and a textual input sequence $x$, RAG aims to generate an output sequence $y$, conditioned on both $D$ and $x$. To create a coherent and consistent representation of the retrieved evidence set $D$, \MODEL condenses it into a succinct context $c$. Our objective is to find the optimal representation of $c$ for the downstream performance. Thus, the question answering task is formalized as 
% \begin{gather*}
%     P_{t}(y \mid x,D) = P_{t}(y \mid x, \hat{c}),  \\
%     \hat{c}=\arg\max_{c}\MODEL(c \mid x,[d_1,d_2,\ldots, d_k]),
% \end{gather*}
% where $P_{t}$ is the target model’s probability distribution, $x$ is the question input, and $[\cdot,\cdot]$ is concatenation.

\subsection{RAG with Evidence Compression}
\label{sec:preliminary}
%Previous research has primarily concentrated on question-focused compression; however, this approach may lead to suboptimal performance in downstream tasks due to the compressed context's potential unfamiliarity with the target model employed. This unfamiliarity arises from discrepancies in internal knowledge and prompt preferences between the compression module and the target model. To address this, \MODEL aims to generate a compressed context that is more aligned with the target model's preferences, thereby improving its performance on downstream tasks. 
% \MODEL achieves this goal by ensembling the logits of two LMs at decoding time, a \emph{compression model} and the \emph{target model}.
%We hereby introduce compression-based retrieval augmented generation.

Given a set of $k$ retrieved evidence snippets $D=\{d_1,d_2,\ldots, d_k\}$ and a textual input sequence $x$, RAG aims to generate an output sequence $y$, conditioned on both $D$ and $x$. However, RAG directly utilizes $D$ which often contains irrelevant information to $x$, potentially confusing the target model in downstream tasks \citep{shi2023large}. %standard RAG often fails to produce a coherent and consistent representation of $D$ for the target LM. 
Thus, we use an additional compression model to condense $D$ into a concise and input-relevant context $c$, which is then used in place of $D$ during the downstream generation process. Thus, the RAG with evidence compression is formalized as: 
\begin{gather*}
    y^{*}=\arg\max_{y}P_{\text{tar}}(y \mid x, \hat{c}),  \\
    \hat{c}=P_{\text{comp}}(c \mid x,[d_1,d_2,\ldots, d_k]),
\end{gather*}
where $y^*$ is the final output sequence, $[\cdot,\cdot]$ denotes concatenation, and $P_{\text{tar}}$ and $P_{\text{comp}}$ represent the probability distributions of the target and compression models, respectively. In this work, we consider any natural language prompting tasks, such as open-domain QA tasks, where $x$ represents the input prompt (also known as the query in QA tasks) and $y^*$ denotes the output sequence.

The compression model's objective is to produce a concise yet informative summary $c$ of the evidential documents $D$ that captures the essential information relevant to the input query $x$. We use an unsupervised approach, where the model is instructed to generate a query-relevant summary of $D$ in a zero-shot manner using an evidence compression instruction prompt, denoted as $I_{comp}$, such as the one below: 
\begin{tcolorbox}[colback=white, colframe=gray!75!gray, boxrule=0.5pt, fonttitle=\small, width=\linewidth, title=Evidence 
Compression Instruction]
\small
Given a question and multiple document snippets, generate one summarized context that is helpful to answer the question.
\end{tcolorbox}
Specifically, the evidence compression is done in an auto-regressive way formalized as, 
\begin{align*}
P_{\text{comp}}(c \mid \mathcal{C}_{\text{comp}}) = \prod_{i=1}^{|c|} P_{\text{comp}}(c_i \mid \mathcal{C}_{\text{comp}}, c_{<i}), 
\end{align*}

where $\mathcal{C}$ denotes the input prompt, constructed by stringifying $\{I_{\text{comp}}, x, D\}$ using a predefined prompt template and $|c|$ is the length of the summary $c$. 

% \subsubsection{Target Model}
% During the context compression, we employ the target model, which is eventually used to generate the final output. The target model is directed to generate a context $c$ that would be helpful in answering the question $x$ without referencing the evidence set. This is also done in zero-shot using a context-generation instruction prompt $I_{gen}$ such as:
% \begin{quote}
% “You are an expert in context generation. Given a question, generate a context that is helpful to answer the question."
% \end{quote}
% The context generation is also performed in an auto-regressive fashion, represented as:
% \begin{align*}
% P_{t}(c \mid I_{gen}, x)=\prod_{i=1}^{|c|}P_{t}(c_i|I_{gen},x,c_{<i})
% \end{align*}
% where $P_{t}$ represents the probability distribution of the target model over the vocabulary, and $|c|$ denotes the length of the generated context $c$.

\subsection{Ensemble Decoding for \MODEL}
\label{sec:monocomp}
%\qin{A smooth intro here. For example: ``To introduce the constraint from the target model for a better compression \{briefly mention our motivation here\}, ...'', the expression here need to be consistent with previous sections.}

Simple compression techniques might lead to subpar performance in downstream tasks because the compressed evidence may not be familiar to the target model. To better align the context to the target model, \MODEL proactively composes it to lower the target model's perplexity by introducing a constraint in decoding space from the target model during the evidence compression. \MODEL achieves this goal through ensemble decoding, which involves a multiplicative ensemble of two LMs—compression model and target model—at each decoding step.

% \subsubsection{Target Model}
%During the context compression, we employ the target model, which is eventually used to generate the final output. 
Specifically, the target model is instructed to generate a context $c$ that would be helpful in answering the question $x$ without referencing the evidence set. This is also done in zero-shot using a context generation instruction prompt $I_{gen}$ such as:
\begin{tcolorbox}[colback=white, colframe=gray!75!gray, boxrule=0.5pt, fonttitle=\small, width=\linewidth, title=Context Generation Instruction]
\small
Given a question, generate a context that is helpful to answer the question.
\end{tcolorbox}
The context generation is also performed in an auto-regressive fashion, represented as:
\begin{align*}
P_{\text{tar}}(c \mid \mathcal{C}_{\text{gen}})=\prod_{i=1}^{|c|}P_{\text{tar}}(c_i|\mathcal{C}_{\text{gen}},c_{<i}),
\end{align*}
where %$P_{t}$ represents the probability distribution of the target model over the vocabulary, and 
$\mathcal{C}_{\text{gen}}$ denotes the input prompt constructed using $\{I_{\text{gen}}, x\}$\footnote{We provide the prompt templates for evidence compression and context generation in \Cref{tab:mono_sys_prompts}.} and
$|c|$ denotes the length of the generated context $c$.%\footnote{We slightly abuse the notation $c$ to denote both the summary and the generated context, as they serve as relevant context to the question.}

Once the compression model and the target model generate their respective probability distributions for the next token, the subsequent token is chosen by maximizing the weighted sum of the log probabilities from both models. The selected token is the continuation of the previously generated text aligned with their objectives. This process is formalized as follows:
\begin{equation*}
    \begin{aligned}
    c_i&=\arg\max_{c'_i, c''_i \in V} (\alpha \cdot \log P_{\text{tar}}(c'_i \mid \mathcal{C}_{\text{gen}},c_{<i}) \\
    &+ (1-\alpha) \cdot \log P_{\text{comp}}(c''_i \mid \mathcal{C}_{\text{comp}},c_{<i})),
    \end{aligned}
\end{equation*}
where $c_i$ is the subsequent token, and $\alpha$ is the ensemble coefficient that weighs between the two probability distributions. We demonstrate how the coefficient $\alpha$ impacts both the perplexity and the downstream performance in \Cref{sec:coef_exp}.

Ensemble decoding proactively shifts the token search space in evidence compression by upweighting those tokens with lower perplexity from the target model's perspective, resulting in a compressed evidence that is more familiar to the target model. Note that since both objectives ultimately share the goal of generating context relevant to the question, combining the logits ensures alignment with this ultimate goal.

In addition, ensemble decoding enables \MODEL to seamlessly integrate both retrieval knowledge from the external evidence set and the target model's parametric knowledge. Specifically, \MODEL selects the $\arg\max$ token from the target model only when the token's probability is higher than that of the compression model, demonstrating that \MODEL draws on parametric knowledge only when necessary—potentially when the compression model is uncertain about the next token. This is particularly beneficial for complex tasks like multi-document QA, where the evidence set may not include all the necessary information \citep{mallen2023not}. In such cases, the missing information in compressed evidence can be supplemented by tokens generated from context generation by the target model, which is entirely based on parametric knowledge. We demonstrate in \Cref{sec:para_nonpara} and \Cref{sec:case_study} that \MODEL can incorporate knowledge from both sources effectively, leading to a performance boost compared to compression methods that solely focus on distilling knowledge from the evidence set.

\section{Experimental Settings}
We assess the effectiveness of \MODEL on knowledge-intensive QA tasks. In this section, we delve into the details of the experimental settings.

\subsection{Datasets}
We evaluate \MODEL on five open-domain QA datasets, including two single-document QA datasets, Natural Questions (NQ; \citealt{kwiatkowski2019natural}) and TriviaQA (TQA; \citealt{joshi2017triviaqa}), and three multi-document QA datasets, HotpotQA (HQA; \citealt{yang2018hotpotqa}), 2WikiMultiHopQA (Wiki; \citealt{ho2020constructing}), and MuSiQue (MQ; \citealt{trivedi2022musique}).
%For single-document QA datasets, we use Natural Questions (NQ)\footnote{https://github.com/google-research-datasets/natural-questions/} \citep{kwiatkowski2019natural} and TriviaQA (TQA)\footnote{https://nlp.cs.washington.edu/triviaqa/} \citep{joshi2017triviaqa}. For multi-document QA datasets, we use HotpotQA\footnote{https://hotpotqa.github.io/} \citep{yang2018hotpotqa}, 2WikiMultiHopQA (2WikiMQA)\footnote{https://github.com/Alab-NII/2wikimultihop} \citep{ho2020constructing}, and MuSiQue\footnote{https://github.com/StonyBrookNLP/musique} \citep{trivedi2022musique}. 
Following prior studies \citep{asaiself,xu2024recomp}, we evaluate the performance on the development set of each dataset using two evaluation metrics, %: (1) \textbf{Accuracy (Acc)}, which measures whether the predicted output contains the ground truth answer and (2) \textbf{F1}, which represents the token-level F1-score of the predicted output string and the ground-truth answer string.
Accuracy (Acc) and token-level F1.

\subsection{Implementation Details}
For all the comparison methods, we utilize \texttt{Llama3-8B-Instruct} and \texttt{Mixtral-8x7B- Instruct} as the target model to tackle downstream QA tasks with RAG. For \MODEL and Zero-shot Summarization, we employ two compression models, one for each target model: \textbf{\texttt{Llama3.2-3B-Instruct}} for \texttt{Llama3- 8B-Instruct} target model and \textbf{\texttt{Mistral-7B- Instruct}} for \texttt{Mixtral-8x7B-Instruct} target model.
For each question, we retrieve five documents from 2018 Wikipedia corpus \citep{karpukhin2020dense} using Contriever-MSMARCO \citep{izacard2021unsupervised}, so as to be consistent with previous studies \citep{xu2024recomp,yoon2024compact}.
%that is fine-tuned on MSMARCO \citep{bajaj2016ms}.
%For \MODEL, we use the same LM for both the compression model and the target model. Note that the compression model and the target model can be different, as long as they share the same tokenizer to ensure compatible token logits, as demonstrated in \Cref{sec:add_exp_res}. 
%For \MODEL and Zero-shot Summarization, we employ two compression models for each target model: \texttt{Llama3.2-3B-Instruct} and \texttt{Llama3-8B-Instruct}, (2) \texttt{Mistral-7B-Instruct} and \texttt{Mixtral-8x7B-Instruct}. 
We set ensemble coefficient $\alpha$ of \MODEL to 0.5 by default, for which more analyses are given in \Cref{sec:coef_exp}. The prompts used in the experiment are presented in \Cref{sec:prompt_temp}.

\subsection{Baselines}
We consider the following categories of baselines. (1) \textbf{No Context}: RAG without any context. (2) \textbf{Gold Compression}: RAG using directly relevant evidence from the retrieved documents if they exist. (3) \textbf{Raw Document}: RAG with raw documents that have not undergone any compression. (4) \textbf{Generated Context} \citep{yu2023generate}: RAG with context generated by the same LM as the target model. This is equivalent to \MODEL with $\alpha=1$, as we rely solely on the target model to generate context when $\alpha=1$. (5) \textbf{Reranking-based Methods}: We rerank sentences in the evidence set and choose top-ranked sentences as the context. We utilize two rerankers—Sentence-BERT \citep{reimers2020making} and RECOMP-extractive \citep{xu2024recomp}. (6) \textbf{Compression-based Methods}: We employ four compressors—LongLLMLingua \citep{jiang2023longllmlingua}, RECOMP-abstractive \citep{xu2024recomp}, CompAct \citep{yoon2024compact}, and Zero-shot Summarization. For Zero-shot Summarization, we use the same evidence compression instruction prompt of \MODEL to summarize multiple evidence using the same LM as the target model. This is equivalent to \MODEL with $\alpha=0$, as we depend entirely on the compression model without any intervention from the target model.\footnote{A more detailed explanation of the implementation of the baselines is provided in \Cref{sec:impl_details}.}

\begin{table*}[ht!]
\small
\centering
\begin{tabular}{lccccccccccc}
\toprule
\multirow{2}{*}{\textbf{Methods}}& \multirow{2}{*}{\textbf{Size}} & \multicolumn{2}{c}{\textbf{NQ}} & \multicolumn{2}{c}{\textbf{TQA}} & \multicolumn{2}{c}{\textbf{HQA}} & \multicolumn{2}{c}{\textbf{Wiki}} & \multicolumn{2}{c}{\textbf{MQ}} \\
\cmidrule(r){3-12}
& & \textbf{Acc} & \textbf{F1} & \textbf{Acc} & \textbf{F1} & \textbf{Acc} & \textbf{F1} & \textbf{Acc} & \textbf{F1} & \textbf{Acc} & \textbf{F1} \\
\midrule
\multicolumn{12}{c}{\textit{Llama3-8B-Instruct}} \\
\midrule
Gold Compression&- & - & - & - & - & 42.3 & 51.3 & 35.7 & 40.0 & 10.2 & 17.7 \\
\midrule
No Context&-& 26.9 & 31.9 & 57.2 & 61.2 & 19.1 & 25.5 & 20.5 & 25.0 & 5.4 & 13.0 \\
Raw Document&- & 42.6 & \textbf{47.1} & 67.6 & 70.8  & 30.3 & 38.7 & 22.0 & 26.8 & 8.2 & 15.0 \\
Generated Context&- & 32.3 & 36.6  & 59.7 & 62.4 & 22.7 & 29.7 & 24.8 & 28.7 & 7.6 & 14.8 \\
\midrule
Sentence-BERT&110M& 30.3 & 35.4 & 59.2 & 62.9  & 22.4 & 29.6 & 18.1 & 22.9 & 7.7 & 14.8   \\
RECOMP-extractive&110M$^\dagger$ & 33.7 & 38.1  & 59.4 & 62.8  & 22.5 & 29.8 & 18.0 & 22.4 & 8.1 & 15.5 \\
\midrule
LongLLMLingua&7B$^\dagger$ & 35.4 & 40.9 & 64.8 & 67.6 & 25.9 & 34.7 & 19.2 & 24.2 & 7.7 & 14.4 \\
RECOMP-abstractive&775M$^\dagger$ & 39.3 & 43.3 & 62.9 & 66.1  & 27.0 & 34.8  & 20.5 & 25.0  & 7.3 & 14.8  \\
CompAct&7B$^\dagger$ & 42.3 & 46.1 & 67.0 & 69.7  & 29.8 & 37.5  & 21.4 &  26.6  & 9.2 & 16.9 \\
\midrule
Zero-shot Summarization&3B & 39.4 & 43.2  & 64.2 & 67.1 & 30.1 & 38.5 & 25.7 & 31.1 & 7.7 & 15.3 \\
\textbf{\MODEL}&3B & \textbf{42.8} & 46.8 & \textbf{68.0} & \textbf{70.9} & \textbf{33.0} & \textbf{41.6} & \textbf{29.6} & \textbf{35.2}& \textbf{10.8} & \textbf{19.9}  \\
\midrule
\midrule
\multicolumn{12}{c}{\textit{Mixtral-8x7B-Instruct}} \\
\midrule
Gold Compression&- & - & - &- &  - &  48.2 & 55.1 & 49.9 & 51.9 & 12.9 & 18.6 \\
\midrule
No Context&- & 36.7 & 38.4  & 68.9 & 72.0 & 25.1 & 31.6 &  32.5 & 35.9 & 6.4 & 11.8 \\
Raw Document&- & \textbf{46.3} & 42.1 & 72.1 & 71.1 & 34.0 & 39.0 & 32.9 & 36.3 & 10.1 & 15.6 \\
Generated Context&- & 33.6 & 33.9 & 61.4 & 62.9 & 26.5 & 32.9 & 30.2 & 34.3 & 7.2 & 13.4 \\
\midrule
Sentence-BERT&110M & 36.8 & 36.8 & 67.0 & 68.7 & 28.3 & 34.5 & 32.5 & 36.2 & 9.9 & 15.2 \\
RECOMP-extractive&110M$^\dagger$ & 38.0 & 37.9  & 66.7 & 68.0 &  28.7 & 34.3  & 31.8 & 34.9 & 9.4 & 15.6   \\
\midrule
LongLLMLingua&7B$^\dagger$ & 40.1 & 39.4 & 70.5 & 71.0 & 32.0 & 38.3 & 31.9 & 36.1 & 9.7 & 15.9 \\
RECOMP-abstractive&775M$^\dagger$ & 42.1 & 41.3 & 68.4 & 69.4 & 32.3 & 38.5  & 32.2 & 36.2 & 7.9 & 13.6\\
CompAct&7B$^\dagger$ & 44.1 & 43.4& 70.3 & 71.4 & 35.2 & 41.6 & 35.9 & 39.5 & 11.2 & 16.9 \\
\midrule
Zero-shot Summarization&7B & 42.1 & 40.6 & 65.9 & 67.0 & 31.4 & 38.1 & 28.5 & 32.8 & 8.4 & 13.8 \\
\textbf{\MODEL}&7B & 43.6 & \textbf{44.5} & \textbf{72.6} & \textbf{73.9} & \textbf{36.3} & \textbf{44.4} & \textbf{40.5} & \textbf{45.2} & \textbf{13.4} & \textbf{19.9} \\
\bottomrule
\end{tabular}
\caption{Experimental results on five open-domain QA datasets. \textbf{Size} column represents the size of the compression model used for each method. $^\dagger$ indicates a fully-supervised compression model, where the compressor is trained.}
\label{tab:main}
\vspace{-1em}
\end{table*}
% The best Accuracy and token-level F1 scores for each dataset are in bold.
% The same LMs are employed for both the compression and the target models for \MODEL. 

\section{Experimental Results}
In this section, we compare the overall performance of \MODEL with other baselines across the five datasets (\Cref{sec:main_results}), explore the impact of ensemble coefficient $\alpha$ on performance and perplexity (\Cref{sec:coef_exp}), investigate how effectively \MODEL incorporate parametric and non-parametric knowledge (\Cref{sec:para_nonpara}), and compare the compression rates with other baselines (\Cref{sec:comp_rate}).
\subsection{Main Results}
\label{sec:main_results}
The overall performance of \MODEL and the baselines across the five datasets are presented in \Cref{tab:main}.\footnote{We present additional experimental results using other combinations of compression and target model at \Cref{sec:add_exp_res}.} To start with, the compression-based methods consistently outperform the reranking-based methods, due to the fact that the reranking-based methods are prone to losing more question-relevant information by discarding lower-ranked sentences. 

Next, \MODEL outperforms all other baselines across all the datasets, except for the Gold Compression which is regarded as the upper bound of the performance. It is noteworthy that \MODEL, as a training-free strategy, outperforms all the supervised compression-based baselines that use similar or larger compression models\footnote{We conduct a fair comparison with RECOMP-abstractive by using the same base compression model in \Cref{sec:head2head}.}. This result suggests that knowledge distillation from a larger teacher LM to a smaller compression model may not generalize well, as the context preferences and prior knowledge of the target model and the teacher model are likely to differ. In contrast, the superior performance of \MODEL is attributed to its ability to familiarize evidence with the target model and its effective incorporation of parametric knowledge from ensemble decoding. 
%For a fairer comparison with RECOMP-abstractive which uses a smaller size compression model than \MODEL, we conduct a head-to-head comparison with RECOMP-abstractive using the same base compression model in \Cref{sec:head2head}. 
Moreover, for the MQ dataset, \MODEL even outperforms Gold Compression baseline which can be viewed as a perfect compressor. This demonstrates that explicitly incorporating parametric knowledge from the target model can significantly enhance performance in multi-document QA, even when the context is imperfect.

Finally, given that Zero-shot Summarization corresponds to \MODEL with $\alpha=0$ and Generated Context corresponds to \MODEL with $\alpha=1$, the fact that \MODEL outperforms both baselines highlights its ability to effectively incorporate tokens from both sources—evidence summary and generated context. This results in superior performance compared to relying on one source alone. 

%These experimental analyses also extend to the experiment in \Cref{sec:add_exp_res}, where the compression and target models differ, further emphasizing \MODEL's generalizability across various model combinations.
\begin{figure}[t]
    \centering
    \includegraphics[width=\linewidth]{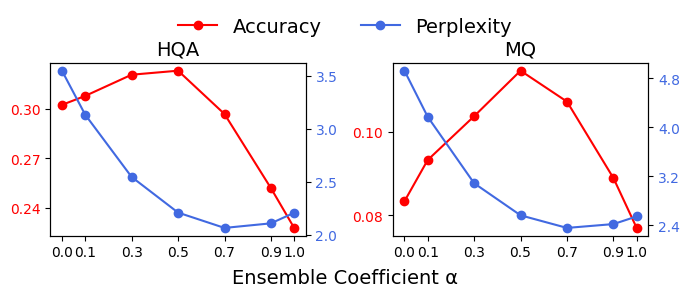}
    \caption{Impact of coefficient $\alpha$ on performance and perplexity when using \texttt{Llama3.2-3B-Instruct} and \texttt{Llama3-8B-Instruct} compression-target pairs.}  
    \label{fig:coef_exp}
\vspace{-1em}
\end{figure}

\subsection{Impact of Ensemble Coefficient on Performance and Perplexity}
\label{sec:coef_exp}
\Cref{fig:coef_exp} illustrates how performance and perplexity change as the ensemble coefficient $\alpha$ is varied across the values when using \texttt{Llama3.2-3B-Instruct} and \texttt{Llama3-8B- Instruct} compression-target pairs on HQA and MQ datasets\footnote{Results for other datasets are included in \Cref{fig:add_coef_exp}.}. We calculate the perplexity of the compressed evidence conditioned on the preceding inputs, i.e. instruction, demonstrations, and the question. For all the datasets, performance is the highest when $\alpha=0.5$, indicating that proactively lowering perplexity by equally weighting both input sources yields the best results. When $\alpha$ is below 0.5, performance improves as the perplexity of compressed evidence decreases, which aligns with the previous works \citep{liu2024monotonic,gonen2023demystifying}. However, when $\alpha$ exceeds 0.5, performance declines as perplexity decreases due to the lack of evidential knowledge during evidence compression. Additionally, when $\alpha$ reaches 0.9 or 1.0, there is a slight rise in the perplexity due to LM's increased uncertainty with limited evidential knowledge. 
% , which aligns with the previous works \citep{liu2024monotonic,gonen2023demystifying}. 
%Interestingly, when $\alpha$ is equal to 0.9 or 1.0, there is a slight increase in perplexity. At high $\alpha$ values, \MODEL is more likely to generate context without referring to external knowledge. As a result, the LM's increased uncertainty when generating context with limited evidential knowledge is likely the cause of the slight rise in perplexity. Results for other datasets are included in \Cref{fig:add_coef_exp}.
% \begin{figure*}[h!]
%     \centering
%     \includegraphics[width=0.6\linewidth]{images/hits_exp.png}
%     \caption{Accuracy of baselines on $Hits=0$ and $Hits=1$ subset of multi-document QA datasets.}  
%     \label{fig:hits_exp}
%     \vspace{-1em}
% \end{figure*}

\begin{figure}[t]
     \centering
     \includegraphics[width=\linewidth]{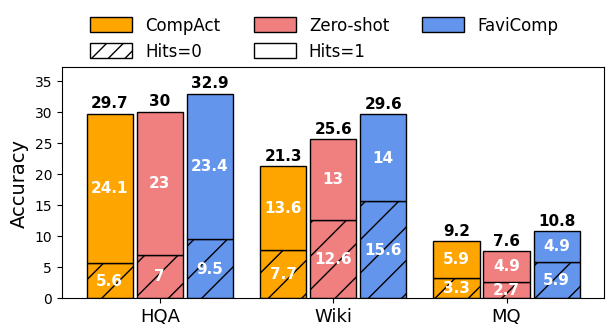}
    \caption{Accuracy of baselines methods on $\mathrm{Hits=0}$ and $\mathrm{Hits=1}$ subset of multi-document QA datasets.}
    \label{fig:hits_exp}
    \vspace{-1em}
\end{figure}

\begin{table}[t]
\centering
\begin{adjustbox}{width=\linewidth}
\begin{tabular}{lccccc}
\toprule
\textbf{Methods} & \textbf{NQ} & \textbf{TQA} & \textbf{HQA} & \textbf{Wiki} & \textbf{MQ} \\
\midrule
Generated Context & 36.6 & 62.4 &  29.7 & 28.7 & 14.8 \\
Zero-shot Summarization & 43.2 & 67.1 &  38.5 & 31.1 & 15.3 \\
Concatenation & 42.5 & 66.7 & 36.5 & 29.0 & 15.6 \\
\MODEL & \textbf{46.8} & \textbf{70.9} & \textbf{41.6} & \textbf{35.2} & \textbf{19.9} \\
\bottomrule
\end{tabular}
\end{adjustbox}
\caption{Performance (F1) comparison against concatenation of parametric and non-parametric knowledge.}
\label{tab:concat}
\vspace{-1em}
\end{table}

\subsection{Integration of Parametric and Non-parametric Knowledge}
\label{sec:para_nonpara}
The effective integration of parametric and non-parametric knowledge is crucial for complex tasks such as multi-document QA, where the evidence set may not contain all the necessary information. To this end, we evaluate how effectively \MODEL incorporates parametric knowledge from the target model and non-parametric knowledge from the compression model on the multi-document QA datasets. We begin by dividing the test samples of each dataset into evidence-relevant and evidence-irrelevant subsets, using the $\mathrm{Hits}$ metric. The $\mathrm{Hits}$ metric is set to 1 (evidence-relevant) if the retrieved evidence set contains the correct answer, and 0 (evidence-irrelevant) if it does not. We then assess the downstream performance of each subset. The underlying intuition is that if a method performs better on the evidence-relevant subset, it suggests that the method is more effectively utilizing the provided evidential knowledge. Conversely, if a method excels on the evidence-irrelevant subset, it indicates that the method is more effectively leveraging parametric knowledge without relying on potentially irrelevant evidence. 

As shown in \Cref{fig:hits_exp}, we compare the accuracy of \MODEL with \texttt{Llama3.2-3B-Instruct} and \texttt{Llama3-8B-Instruct} compression-target pairs on $\mathrm{Hits=0}$ and $\mathrm{Hits=1}$ subsets with the top-performing baselines, Zero-shot Summarization and CompAct\footnote{We provide results of \MODEL on various alpha values in \Cref{sec:hits_exp_alpha}}. \MODEL outperforms other baselines in the $\mathrm{Hits=0}$ subset while performing comparably with others in the $\mathrm{Hits=1}$ subset. This proves that \MODEL effectively relies on parametric knowledge rather than evidential knowledge when faced with irrelevant evidence, while maintaining similar effectiveness in utilizing evidential knowledge when relevant evidence is present. 

% Interestingly, even though CompAct generally performs better on the $\mathrm{Hits=1}$ subset compared to Zero-shot Summarization, it underperforms relative to Zero-shot Summarization on the $\mathrm{Hits=0}$ subset. This suggests that the training may have been biased towards utilizing solely evidential knowledge, rather than effectively leveraging both sources in synergy.

In addition, we conduct another experiment to demonstrate \MODEL's superior ability to synergize two sources of knowledge. We compare it against a straightforward approach that concatenates parametric and non-parametric knowledge as context for downstream generation. Specifically, we concatenate the compressed evidence from the Zero-shot Summarization with the generated context from the Generated Context and use this concatenated context for evaluation. The results, shown in \Cref{tab:concat}, reveal that simple concatenation underperforms compared to the Zero-shot Summarization baseline. This suggests that naively merging non-parametric and parametric knowledge in-context can be less effective than relying solely on non-parametric knowledge. In contrast, \MODEL effectively integrates both knowledge sources during compression, leveraging their synergy to achieve superior performance.

\subsection{Compression Rate Comparisons}
\label{sec:comp_rate}
Since one of the functionalities of evidence compression in RAG is to reduce the number of tokens from the evidence set, we report the compression rate of \MODEL with \texttt{Llama3.2-3B-Instruct} and \texttt{Llama3-8B-Instruct} compression-target pairs in \Cref{tab:compression}. We compute the compression rate as $\tfrac{\textit{\# of tokens in retrieved documents}}{\textit{\# of tokens in compressed documents}}$. Overall, RECOMP-abstractive and \MODEL consistently score the highest compression rates. %Reranking-based methods achieve high compression since they only select one or two sentences that may contain the answer to the question, but the information loss is more significant compared to other methods. 
RECOMP-abstractive exhibits high compression rates because the compression model is trained to output an empty string when no relevant evidence is found, which is often the case in multi-document QA datasets. \MODEL compresses the evidence to make it familiar to the target model by lowering its perplexity at each decoding step, typically resulting in a shorter context. Notably, when compared to Zero-shot Summarization, which is equivalent to \MODEL with $\alpha=0$, \MODEL consistently achieves higher compression rates. This demonstrates that the ensemble decoding strategy, combining token logits from both evidence compression and context generation, leads to greater compression efficiency. 
\begin{table}[t]
\centering
\begin{adjustbox}{width=\linewidth}
\begin{tabular}{lccccc}
\toprule
\textbf{Methods} & \textbf{NQ} & \textbf{TQA} & \textbf{HQA} & \textbf{Wiki} & \textbf{MQ} \\
\midrule
LongLLMlingua & 1.87 & 1.84 & 1.83 & 1.83 & 1.83 \\
RECOMP-abstractive & \textbf{17.96} & \underline{17.79} & \underline{19.72} & \textbf{32.06} & \textbf{32.05} \\
CompAct & 8.85 & 8.92 & 9.45 & 10.71 & 8.96 \\
Zero-shot Summarization & 14.12 & 17.12 & 18.75 & 21.39 & 16.19 \\
\textbf{\MODEL} & \underline{16.43} & \textbf{22.40} & \textbf{22.55} & \underline{23.10} & \underline{18.95} \\
\bottomrule
\end{tabular}
\end{adjustbox}
\caption{Compression rates of the baselines and \MODEL.}
\label{tab:compression}
\vspace{-1em}
\end{table}

\begin{table*}[t]
\small
\centering
\resizebox{\textwidth}{!}{
\begin{tabular}
{m{0.15\linewidth} m{0.55\linewidth} m{0.2\linewidth} m{0.1\linewidth}}
\toprule
\multicolumn{4}{c}{\textbf{Question}: This film is an adaption of a Jacques Offenbach's opera that was written by a Hungarian British screenwriter?} \\
\midrule
\centering \textbf{Methods} & \centering \textbf{(Compressed) Evidence} & \centering \textbf{Prediction} & \textbf{Perplexity} \\
\midrule
Raw Document & The Tales of Hoffmann is a 1951 British Technicolor film adaptation of Jacques Offenbach's opera ``The Tales of Hoffmann", written, produced and directed by the team of Michael Powell and Emeric Pressburger ... & Emeric Pressburger \redcross & 12.429\\
\midrule
Zero-shot 

Summarization & The 1951 film ``The Tales of Hoffmann" is an adaptation of Jacques Offenbach's opera, written, produced, and directed by Michael Powell and Emeric Pressburger. & Emeric Pressburger \redcross & 2.298
\\
\midrule
\MODEL & The 195\pink{1} film ``The Tales of Hoffmann" is an adaptation of Jacques Offenbach's opera, written\skyblue{ by}\pink{ Emeric} Pressburger,\skyblue{ a} Hungarian\pink{-B}ritish screenwriter\pink{,} and\purple{ directed} by Michael Powell and Emeric Pressburger. & The Tales of 

Hoffmann \greencheck & 1.959
\\
\midrule
\midrule
\multicolumn{4}{c}{\textbf{Question}: Which magazine was first published earlier, The Chronicle of Philanthropy or Skeptic?} \\
\midrule
Raw Document & The Chronicle of Philanthropy is a magazine that covers the nonprofit world. ... It was founded in 1988 by editor Phil ... First published as a newsletter in 1987, "Philanthropy" became a glossy magazine in 1996. & Philanthropy \redcross & 4.856 \\
\midrule
Zero-shot 

Summarization & The Chronicle of Philanthropy was founded in 1988, while Philanthropy magazine was first published as a newsletter in 1987 and became a glossy magazine in 1996. & Philanthropy 

magazine \redcross & 3.196
\\
\midrule
\MODEL & The Chronicle of Philanthropy was\skyblue{ first} published in 1988, while\skyblue{ Ske}ptic was\skyblue{ first} published\skyblue{ in} 199\skyblue{2}. & The Chronicle of 

Philanthropy \greencheck & 1.345
\\
\bottomrule
\end{tabular}
}
\caption{Case study of evidence compression: \MODEL vs. Raw Document and Zero-shot Summarization. For \MODEL, the colors \pink{red} and \skyblue{blue} highlight tokens that are the $\arg\max$ of the compression model and the target model, respectively. \purple{Purple} indicates a token that is the $\arg\max$ of neither model. Tokens with no coloring represent those that are the $\arg\max$ of both models.}
\label{tab:case_study}
\vspace{-1em}
\end{table*}

\section{Case Study}\label{sec:case_study}

\Cref{tab:case_study} presents two examples from HQA to illustrate how \MODEL effectively familiarizes evidence while seamlessly integrating both parametric and non-parametric knowledge during evidence compression. We compare its output with Raw Document, which does not apply any compression, and Zero-shot Summarization. 

In both examples, Raw Document fails to produce the correct answer, even though the evidence contains the necessary information, highlighting the need for effective evidence compression. In the first example, while the difference between the compressed evidence from Zero-shot Summarization and \MODEL appears subtle, \MODEL delivers the correct answer with a lower perplexity in compression, underscoring the significance of evidence familiarization. The second example highlights the importance of parametric knowledge when the retrieved evidence set lacks complete information. Since the evidence set does not mention "Skeptic", Zero-shot Summarization introduces irrelevant information ("Philanthropy magazine"), ultimately leading to an incorrect answer. In contrast, \MODEL integrates parametric knowledge about "Skeptic" and incorporates it into the evidence compression. Notably, \MODEL selects the $\arg\max$ token from the target model only when the token's probability is higher than that of the compression model, demonstrating that \MODEL draws on parametric knowledge only when necessary—potentially when the compression model is uncertain about the next token.

\section{Related Works}
\stitle{Evidence Compression for RAG}
%\ben{Can we rename this to something like Evidence selection? There are several works (see the two I added) does not use compression, but decomposition-based IR).}
% Standard RAG retrieves textual evidence related to the prompt from the external corpora or knowledge bases and incorporates it as a part of the input to the LM \citep{lewis2020retrieval,izacard-grave-2021-leveraging,guu2020retrieval}. However, retrieved evidence pieces may contain inconsistent or irrelevant information to the question, potentially confusing the target model in downstream tasks \citep{shi2023large}. To tackle this problem, traditional approaches aim to rerank the textual evidence based on its relevance to the question and then select a top-ranked subset to include as part of the input to the LM \citep{nogueira2020document,zhuang2023rankt5}. However, this approach loses more question-relevant information by discarding lower-ranked sentences. 
Recent efforts on evidence compression seek to compress retrieved evidence pieces to filter out unnecessary information and retain only the essential context \cite{wang2023learning,li2024refiner,ke2024bridging,xu2024recomp,yoon2024compact}. %\citet{wang2023learning} filter query-relevant context using relevance metrics and \citet{li2024refiner} extract query-relevant information and restructure them to form a consistent context. \citet{ke2024bridging} trains a seq2seq bridge model using supervised and reinforcement learning to optimize the connection between the retriever and the LLM. 
Most recently, %\citet{jiang2023longllmlingua} and \citet{cao-etal-2024-retaining} conduct token-level or embedding-based compression to preserve only the query-relevant information using a trained compressor. 
\citet{xu2024recomp} and \citet{yoon2024compact} train a compression model to generate an abstractive summary of the documents by distilling knowledge from larger language models. 

While these methods are successful to some extent, they often achieve suboptimal performance because of the discrepancy between the compression model and the target model, leading unfamiliarity of the context.
In contrast, \MODEL proactively compresses the evidence pieces in a way to lower the target model's perplexity using an ensemble decoding technique without any training, thereby improving the downstream performance.

\stitle{Parametric and Non-parametric Knowledge in RAG}
% Recent studies have investigated the phenomena of LM's utilization of parametric and non-parametric knowledge in RAG scenarios \citep{longpre2021entity,wadhwa2024rags,wu2024clasheval,zhang2024evaluating}. \cite{longpre2021entity} examines how LMs use both types of knowledge when there are entity-based conflicts between them. \cite{wadhwa2024rags} demonstrate that LMs often have a strong tendency to rely solely on context information to answer questions. \cite{wu2024clasheval} further shows that this tendency is particularly pronounced when LMs are less confident in their initial response or when the retrieved documents are more realistic. 
%While there have been studies on the phenomena of LM's utilization of both parametric and non-parametric knowledge sources, there is a lack of research focused on effectively synergizing both sources. 
There has been a lack of research focused on effectively combining both sources. 
A few of these efforts introduce counterfactual augmentation \citep{longpre2021entity,fang2024getting,zhang2024evaluating} and causal intervention \citep{zhou2023context,wang2023causal} to mitigate knowledge conflict, which, however, requires explicitly knowing the features of the input that causes such conflict.
\citet{zhang2023merging} seek to address this issue by incorporating LM-generated context into the LM's input along with the retrieved documents, thereby integrating both sources of knowledge. 
However, merely concatenating both contexts is a suboptimal solution, as LMs may still show bias toward one source over the other when generating responses \citep{longpre2021entity,wu2024clasheval}. To address this, \MODEL employs ensemble decoding during the evidence compression, ensuring that both types of knowledge are seamlessly fused together to create a consistent context.

\stitle{Constrained Decoding}
Constrained decoding has been previously proposed in text generation tasks for various purposes, including optimizing prompts \citep{liu2024monotonic}, enhancing plausibility \citep{li2023contrastive} or controllability \citep{meng2022controllable,huang2023affective}, and reducing hallucination \citep{shi2024trusting}. 
%Contrastive Decoding \citep{li2023contrastive} enforces a plausibility constraint during generation by inducing the difference in token log-probabilities between expert and amateur LMs. 
%Context-aware Decoding \citep{shi2024trusting} uses contrastive decoding to amplify the probability differences between outputs with and without evidence, encouraging the LM to prioritize the evidential knowledge. 
Our work is closely connected with the method by
\citet{liu2024monotonic} which employs ensemble decoding to paraphrase prompts to enhance zero-shot LM prompting and generalization. 
%While similar to our method—both use ensemble decoding  to improve downstream performance—t
Their approach focuses on the robustness and generalizability of instruction prompts for tasks without retrieval augmentation. 
In contrast, our approach compresses externally retrieved evidence while integrating parametric knowledge during compression, specifically targeting knowledge-intensive tasks that require balancing both evidential and parametric knowledge.

\section{Conclusion}
In this study, we introduce \MODEL, a training-free, inference-time evidence compression method designed to enhance RAG performance by consolidating retrieved evidence set to be more familiar to the target model, while seamlessly integrating parametric knowledge. %By leveraging ensemble decoding, \MODEL compresses the retrieved evidence to make it more favorable to the target model. Moreover, \MODEL effectively balances the target model's parametric knowledge and the retrieved knowledge, improving performance on complex tasks where the retrieved evidence set may not contain all the necessary information. 
%\MODEL employs ensemble decoding to familiarize compressed evidence to the target model while balancing parametric and retrieved knowledge, enhancing performance on complex tasks with incomplete evidence.
Our extensive experiments validate the effectiveness of \MODEL on open-domain QA tasks, showing significant improvements over recent evidence compression baselines in multiple datasets. Additionally, \MODEL's model-agnostic nature allows it to be incorporated into various RAG workflows at inference time, making it a versatile tool for enhancing LMs in complex tasks.

\section*{Acknowledgment}

We appreciate the reviewers for their insightful
comments and suggestions.
This work was partly supported by the Amazon Nova Trusted AI Prize, the NSF of the United States Grants ITE 2333736 and OAC 2531126, and the DARPA FoundSci Grant HR00112490370.

\section*{Limitations}

Although \MODEL exhibits superior performance in RAG compared to the recent evidence compression baselines, it has some limitations. (1) \MODEL consumes approximately twice as much computation compared to methods that only use a compression model since it needs two inferences (compression and target model) during the ensemble decoding. However, it is a training-free strategy that can be easily plugged into any RAG application. We provide insights on the tradeoff between latency and performance in \Cref{sec:latency}. (2) Ensemble decoding requires the compression and target model to share the same vocabulary and tokenizer, which can limit the range of compatible models. Nonetheless, recent studies, such as \citet{gu2024chared}, have introduced techniques to enable model-agnostic ensemble decoding. This implies that there will be a potential direction of incorporating model-agnostic ensemble decoding with our framework to enable more flexible integration of various models, which we leave as future work. 

\section*{Ethics Statement}
This work follows the ACL Code of Ethics. We believe no potential risk is directly associated with the presented work.

% Bibliography entries for the entire Anthology, followed by custom entries
%\bibliography{anthology,custom}
% Custom bibliography entries only
\bibliography{custom}

\appendix
\begin{table*}[ht]
\small
\centering

\begin{tabular}{lcccccccc}
\toprule
\multirow{2}{*}{\textbf{Methods}}&\multirow{2}{*}{\textbf{Train}}&\multirow{2}{*}{\textbf{Compression Model}}& \multicolumn{2}{c}{\textbf{NQ}} & \multicolumn{2}{c}{\textbf{TQA}} & \multicolumn{2}{c}{\textbf{HQA}} \\
\cmidrule(r){4-9}
& & & \textbf{Acc} & \textbf{F1} & \textbf{Acc} & \textbf{F1} & \textbf{Acc} & \textbf{F1} \\
\midrule
RECOMP-abstractive & O&  T5-large & 38.0 & 37.8 & 62.1 & 65.0 & 27.4 & 34.3 \\ 
RECOMP-abstractive &O &  Mistral-7B-Instruct-v0.3 & 38.3 & 38.2 & 63.0 & 65.4 & 29.5 & 36.6\\ 
FaviComp & X & Mistral-7B-Instruct-v0.3 & 40.3 & 40.4 & 65.9 & 68.9 & 32.0 & 40.5 \\ 
\bottomrule
\end{tabular}

\caption{Head-to-head comparison results with RECOMP}
\label{tab:head2head_recomp}
\end{table*}

\begin{table}[h]
\small
\centering
\begin{tabular}{lccccc}
\toprule
\textbf{Dataset} & \textbf{NQ} & \textbf{TQA} & \textbf{HQA} & \textbf{Wiki} & \textbf{MQ} \\
\midrule
\textbf{\# of Samples} & 3610 & 11313 & 7405 & 12576 & 4834 \\
\bottomrule
\end{tabular}
\caption{Number of samples in each dataset.}
\label{tab:dataset_stats}
\end{table}
\section{Implementation Details}
\label{sec:impl_details}
\subsection{Generation Configuration} For all the baselines and \MODEL, we use default temperature and top-p values of the compression model during evidence compression and fix the temperature of the target model to 1.0 during evaluation. 
\subsection{Dataset Statistics} We provide the statistics of the evaluation dataset utilized in our experiments in \Cref{tab:dataset_stats}.  
\subsection{Implementation Details of Baselines}
(1) \textbf{Gold Compression}: We implement the Gold Compression baseline following the approach outlined by \cite{yoon2024compact}. We evaluate only on HQA, Wiki, and MQ, as these datasets contain gold documents. We first identify the presence of any gold documents in the retrieved documents. If found, we use the documents as the context. If none of the retrieved documents are identified as gold, we utilize the entire set of retrieved documents as the context for the evaluation. To identify the gold documents within the retrieved documents, we compare each gold document with the retrieved ones. If 50\% or more of the content matches, we classify it as a gold document. This approach is necessary because the documents are chunked, and the retrieved documents may not exactly match the gold documents. 
\\
(2) \textbf{Generated Context}: We use the context generation prompt in \Cref{tab:mono_sys_prompts} to generate the context.
\\
(3) \textbf{Zero-shot Summarization}: We use the evidence compression prompt in \Cref{tab:mono_sys_prompts} to compress the retrieved documents. 
\\
(4) \textbf{RECOMP-extractive}: We utilize the same Contriever models trained by the authors for each dataset, to encode both the question and the sentences in the evidence set. For Wiki and MQ, since there are no fine-tuned models available, we use the Contriever fine-tuned on HQA. Following the original paper, we select one sentence as the context for NQ and TQA, whereas for the other datasets, we utilize two sentences.
\\
(5) \textbf{RECOMP-abstractive}: Similar to RECOMP-extractive, we use the same T5-large models trained by the authors for each dataset to compress the retrieved evidence. For the Wiki and MQ, we employ the T5-large model fine-tuned on HQA.
\\
(6) \textbf{LongLLMLingua}: We use \texttt{Llama2-7B}\footnote{https://huggingface.co/NousResearch/Llama-2-7b-hf} trained by the authors as the prompt compressor model. We use the default hyperparameters in the original paper, where the dynamic context compression rate is set to 0.3, and the maximum compression rate is set to 0.5. 
\\
(7) \textbf{CompAct}: We use the same \texttt{Mistral-7B- Instruct}\footnote{https://huggingface.co/cwyoon99/CompAct-7b} model instruction-tuned by the authors for evidence compression. The number of documents per segment is set to 5 with 1 iteration.

\section{Additional Experiment Results}
\subsection{Other Compression and Target Models}
\label{sec:add_exp_res}
We conduct an experiment where we use \texttt{Llama3 -8B-Instruct} and \texttt{Mistral-7B-Instruct} for both compression and target models. The result in \Cref{tab:appe_result} demonstrates that \MODEL outperforms all other baselines, supplementing the effectiveness shown in \Cref{sec:main_results}.

\subsection{Head-to-Head Comparison with RECOMP-abstractive}
\label{sec:head2head}
Since the lower performance of RECOMP-abstractive might possibly be due to the use of smaller base model for compression (\texttt{T5-large}), we conduct a head-to-head experiment on \MODEL and RECOMP-abstractive by using the same base compression model. We construct training data on NQ, TQA, and HQA according to \citet{xu2024recomp} and finetune \texttt{Mistral-7B-Instruct} on each of the training data. We train for 7 epochs using LoRA with Adam optimizer with a learning rate of 2e-6 and a batch size of 64. We present the evaluation results in \Cref{tab:head2head_recomp}. Even though using larger base model for compression enhances the performance of RECOMP-abstractive to some extent, it still underperforms compared to training-free \MODEL. This underscores that the familiarization during evidence compression and integration of parametric and non-parametric knowledge are more helpful to the downstream generation than relying on a trained model for evidence compression. 

\subsection{Performance of $\mathrm{Hits=0}$ and $\mathrm{Hits=1}$ on Varying Alpha Values}
\label{sec:hits_exp_alpha}
\begin{figure}[t]
     \centering
     \begin{subfigure}[b]{\linewidth}
         \centering
         \includegraphics[width=\linewidth]{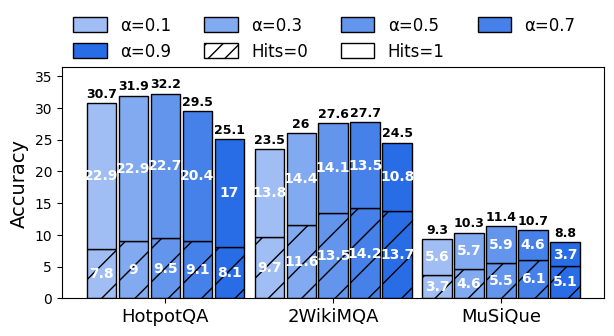}
         \label{fig:hits_exp_alphas}
     \end{subfigure}
    \caption{Accuracy of \MODEL with various $\alpha$ values on $\mathrm{Hits=0}$ and $\mathrm{Hits=1}$ subset of multi-document QA datasets.}
    \label{fig:add_hits_exp}
\end{figure}

We evaluate \MODEL's performance on evidence-relevant ($\mathrm{Hits=1}$) and evidence-irrelevant ($\mathrm{Hits=0}$) subsets by varying $\alpha$ values. \Cref{fig:add_hits_exp} shows that $\alpha=0.5$ or $\alpha=0.7$ performs the best on the $\mathrm{Hits=0}$ subset, while performance declines as $\alpha$ deviates further from the value. This pattern in the $\mathrm{Hits=0}$ subset mirrors the overall performance trend, suggesting that appropriately utilizing parametric knowledge when the evidence is irrelevant is crucial to the overall performance. In the $\mathrm{Hits=1}$ subset, performance remains consistent for $\alpha$ values up to 0.5 but decreases significantly when $\alpha$ exceeds 0.5 due to the diminished utilization of the relevant evidential context.

\begin{table*}[ht]
\small
\centering
\begin{tabular}{lcccc}
\toprule
\multirow{2}{*}{\textbf{Methods}} & \multirow{2}{*}{\textbf{Compression Model}} & \multirow{2}{*}{\textbf{Avg latency per sample (s)}} & \multicolumn{2}{c}{\textbf{Performance}} \\
\cmidrule(r){4-5}
 & & & \textbf{Acc} & \textbf{F1} \\
\midrule
RECOMP-abstractive & T5-large & 0.22 & 39.3 & 43.3 \\ 
CompAct & Mistral-7B-Instruct-v0.2 & 8.72 & 42.3 & 46.1\\ 
Zero-shot Summarization & Llama-3.2-3B-Instruct & 3.99 & 39.4 & 43.2\\ 
\MODEL & Llama-3.2-3B-Instruct & 6.43 & 42.8 & 46.8\\ 
\bottomrule
\end{tabular}

\caption{Latency and of the baselines and \MODEL}
\label{tab:latency}
\end{table*}

\subsection{Latency Ablation Study} 
\label{sec:latency}
\Cref{tab:latency} shows the latency of our method along with other major baselines to provide insights on the trade-offs between accuracy and latency. We used Llama-3-8B-Instruct as the target model and tested on NQ dataset for the experiment. Although there are trade-offs between latency and accuracy across all methods, training-free FaviComp demonstrates lower latency while achieving higher accuracy than CompAct, which is the supervised baseline that previously achieved SOTA performance.

\section{Prompt Templates}
\label{sec:prompt_temp}

\begin{figure}[ht]
\begin{tcolorbox}[colback=white, colframe=gray!75!gray, boxrule=0.5pt, title=Evaluation Prompt Template]
\texttt{\{System Prompt\}} \\
\texttt{\{Demonstrations\}} \\
Question: \texttt{\{Question\}} \\
Context: \texttt{\{Context\}} \\
Answer:
\end{tcolorbox}
\caption{Evaluation Prompt Template.}
\label{fig:eval_prompt_temp}
\end{figure}

\subsection{Evaluation}
The evaluation prompt template is shown in \Cref{fig:eval_prompt_temp}. For all the evaluations throughout the experiment, we switch the positions of the Question and Context if doing so results in better performance. System prompts and demonstrations used in the evaluations are presented in \Cref{tab:eval_sys_prompts} and \Cref{tab:eval_demos}, respectively. 
\subsection{\MODEL}
The prompt templates for evidence compression and context generation of \MODEL are presented in \Cref{tab:mono_sys_prompts}.

\section{Licenses}
We include the licenses of datasets and models we used in this work.
\\
Dataset Licenses:
\begin{itemize}
  \item NQ: Apache-2.0 
  \item TQA: Apache-2.0 
  \item HQA: CC BY-SA 4.0 
  \item Wiki: Apache-2.0 
  \item MQ: CC-BY-4.0 
\end{itemize}
Model Licenses:
\begin{itemize}
  \item Llama3: Custom License \url{https://www.llama.com/llama3/license/}
  \item Mistral \& Mixtral: Apache-2.0 
\end{itemize}

\begin{table*}[t]
\centering
\small
\begin{tabular}{lcccccccccc}
\toprule
\multirow{2}{*}{\textbf{Methods}} & \multicolumn{2}{c}{\textbf{NQ}} & \multicolumn{2}{c}{\textbf{TQA}} & \multicolumn{2}{c}{\textbf{HQA}} & \multicolumn{2}{c}{\textbf{Wiki}} & \multicolumn{2}{c}{\textbf{MQ}} \\
\cmidrule(r){2-11}
& \textbf{Acc} & \textbf{F1} & \textbf{Acc} & \textbf{F1} & \textbf{Acc} & \textbf{F1} & \textbf{Acc} & \textbf{F1} & \textbf{Acc} & \textbf{F1} \\
\midrule
\multicolumn{11}{c}{\textit{Llama3-8B-Instruct}} \\ 
\midrule
Gold Compression & - & - & - & - & 42.3 & 51.3 & 35.7 & 40.0 & 10.2 & 17.7 \\
\midrule
No Context& 26.9 & 31.9 & 57.2 & 61.2 & 19.1 & 25.5 & 20.5 & 25.0 & 5.4 & 13.0 \\
Raw Document & 42.6 & \textbf{47.1} & 67.6 & 70.8  & 30.3 & 38.7 & 22.0 & 26.8 & 8.2 & 15.0 \\
Generated Context & 32.3 & 36.6  & 59.7 & 62.4 & 22.7 & 29.7 & 24.8 & 28.7 & 7.6 & 14.8 \\
\midrule
Sentence-BERT& 30.3 & 35.4 & 59.2 & 62.9  & 22.4 & 29.6 & 18.1 & 22.9 & 7.7 & 14.8   \\
RECOMP-extractive$^\dagger$ & 33.7 & 38.1  & 59.4 & 62.8  & 22.5 & 29.8 & 18.0 & 22.4 & 8.1 & 15.5 \\
\midrule
LongLLMLingua$^\dagger$ & 35.4 & 40.9 & 64.8 & 67.6 & 25.9 & 34.7 & 19.2 & 24.2 & 7.7 & 14.4 \\
RECOMP-abstractive$^\dagger$ & 39.3 & 43.3 & 62.9 & 66.1  & 27.0 & 34.8  & 20.5 & 25.0  & 7.3 & 14.8  \\
CompAct$^\dagger$ & 42.3 & 46.1 & 67.0 & 69.7  & 29.8 & 37.5  & 21.4 &  26.6  & 9.2 & 16.9 \\
\midrule
Zero-shot Summarization & 41.3 & 45.1& 66.3 & 69.5 & 30.2 & 38.6 & 22.3 & 28.1 & 8.3 & 16.3  \\
\textbf{\MODEL} & 42.3 & 46.6  & \textbf{68.4} & \textbf{71.5}  & 32.3 & 41.0  & 27.6 & 33.6  & \textbf{11.4} & \textbf{20.1} \\
\midrule
\midrule
\multicolumn{11}{c}{\textit{Mistral-7B-Instruct}} \\
\midrule
Gold Document& - & - & - & - & 41.0 & 50.5 & 38.1 & 40.3 & 9.6 & 15.2 \\
\midrule
No Context& 28.1 & 27.5 & 58.8 & 60.9 & 19.7 & 24.8 & 21.9 & 22.8 & 5.2 & 9.7 \\
Raw Document& 40.2 & 39.3 & \textbf{66.2} & 68.6 & 30.3 & 37.2 & 26.6 & 28.5 & 7.5 & 13.1 \\
Generated Context& 30.1 & 31.7 & 57.3 & 60.7 & 23.7 & 30.6 & 25.1 & 29.5 & 7.1 & 12.8 \\
\midrule
Sentence-BERT & 29.8 & 30.1 & 57.8 & 60.7 & 23.8 & 30.3 & 22.9 & 24.7 & 7.5 & 12.3 \\
RECOMP-extractive$^\dagger$ & 31.7 & 32.2 & 57.2 & 60.0 & 24.1 & 30.2 & 23.2 & 24.4 & 7.4 & 12.5 \\
\midrule
LongLLMLingua$^\dagger$ & 34.3 & 36.4 & 63.8 & 66.9 & 27.0 & 34.7 & 25.5 & 28.0 & 7.1 & 13.0 \\
RECOMP-abstractive$^\dagger$ & 38.0 & 37.8 & 62.1 & 65.0 & 27.4 & 34.3 & 25.1 & 27.4 & 6.4 & 12.0 \\
CompAct$^\dagger$ & 38.8 & 38.9 & 65.1 & 67.1 & 30.2 & 37.1 & 24.9 & 27.6 & 8.2 & 13.6 \\
Zero-shot Summarization & 38.4 & 38.2 & 62.3 & 64.8 & 28.2 & 35.2 & 23.2 & 27.1 & 6.8 & 11.8 \\
\textbf{\MODEL}& \textbf{40.3} & \textbf{40.4} & 65.9 & \textbf{68.9} & \textbf{32.0} & \textbf{40.5} & \textbf{29.7} & \textbf{35.1} & \textbf{9.2} & \textbf{15.2} \\
\bottomrule
\end{tabular}
\caption{Additional experimental results. \texttt{Llama3-8B-Instruct} and \texttt{Mistral-7B-Instruct} are used for both compression and target models.}
\label{tab:appe_result}
\end{table*}

\begin{figure*}[ht]
    \centering
    \includegraphics[width=0.7\linewidth]{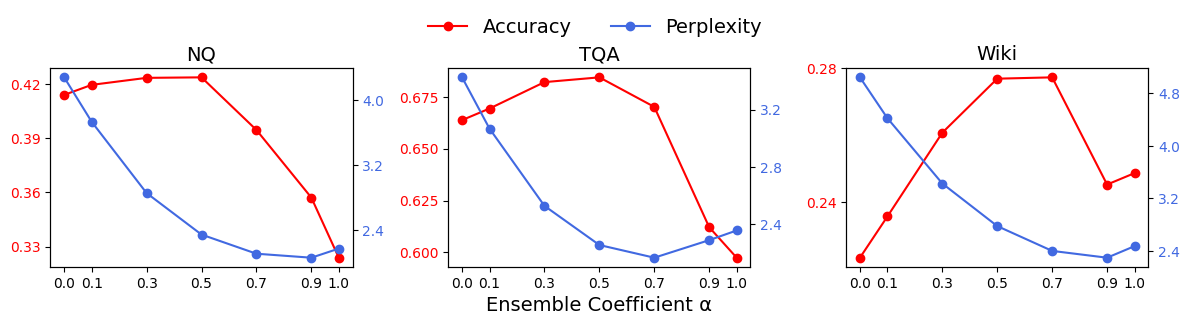}
    \caption{Impact of coefficient $\alpha$ on performance and perplexity for NQ, TQA and Wiki.}  
    \label{fig:add_coef_exp}
    \vspace{-1em}
\end{figure*}
\begin{table*}[ht]
\small
\centering
\begin{tabular}
{m{0.25\linewidth}|m{0.65\linewidth}}
\toprule
\textbf{Target Models} & \textbf{System Prompt}\\
\midrule
Llama-3-8B-Instruct & You are an expert in Question Answering. Your job is to answer questions in 1 to 5 words based on the given context.
\\
\midrule
Mixtral-8x7B-Instruct & You are an expert in Question Answering. Your job is to answer questions in 1 to 5 words based on the given context. Just output the answer as concisely as possible, no other words
\\
\midrule
Mistral-7B-Instruct & You are an expert in Question Answering. Your job is to answer questions in 1 to 5 words based on the given context. Just output the answer as concisely as possible, no other words
\\
\bottomrule
\end{tabular}
\caption{System prompts used in evaluation}
\label{tab:eval_sys_prompts}
\end{table*}

\begin{table*}[ht]
\small
\centering
\begin{tabular}
{m{0.22\linewidth}|m{0.65\linewidth}}
\toprule
\textbf{Instruction} & \textbf{Prompt Template}\\
\midrule
Evidence Compression & You are an expert in summarization. Given a question and multiple document snippets, generate one summarized context that is helpful to answer the question. Just summarize, no other words.

Question: \texttt{\{Question\}}

Documents: \texttt{\{Evidence\}}

Summarized Context: 
\\
\midrule
Context Generation & You are an expert in context generation. Given a question, generate a context that is helpful to answer the question. Just generate the context, no other words.

Question: \texttt{\{Question\}}

Context: 
\\
\bottomrule
\end{tabular}
\caption{Prompt Templates for \MODEL}
\label{tab:mono_sys_prompts}
\end{table*}
\begin{table*}[ht]
\small
\centering
\begin{tabular}
{m{0.1\linewidth}|m{0.8\linewidth}}
\toprule
\textbf{Dataset} & \textbf{Demonstrations}\\
\midrule
NQ & Question: who sings i've got to be me

Answer: Sammy Davis, Jr

Question: who wrote i will follow you into the dark

Answer: Ben Gibbard

Question: who won season 2 of total drama island

Answer: Owen (Scott McCord)

Question: what part of the mammary gland produces milk

Answer: cuboidal cells

Question: when did the golden compass book come out

Answer: 1995
\\
\midrule
TQA & Question: Who sang the theme for the James Bond film ‘Thunderball’?

Answer: Tom Jones

Question: A hendecagon has how many sides?

Answer: Eleven

Question: In the 1968 feature film Chitty Chitty Bang Bang, of what country is Baron Bomburst the tyrant ruler?

Answer: Vulgaria

Question: Artists Chuck Close, Henri-Edmond Cross, John Roy, Georges-Pierre Seurat, Paul Signac, Maximilien Luce and Vincent van Gogh painted in what style?

Answer: Pointillism

Question: What is the study of the relation between the motion of a body and the forces acting on it?

Answer: Dynamics
\\
\midrule
HQA & Question: Which magazine was started first Arthur's Magazine or First for Women?

Answer: Arthur's Magazine

Question: The Oberoi family is part of a hotel company that has a head office in what city?

Answer: Delhi

Question: Musician and satirist Allie Goertz wrote a song about the "The Simpsons" character Milhouse, who Matt Groening named after who?

Answer: President Richard Nixon

Question: Are Jane and First for Women both women's magazines?

Answer: Yes

Question: Were Pavel Urysohn and Leonid Levin known for the same type of work?

Answer: No
\\
\midrule
Wiki & Question: Where was the place of death of Marie Thérèse Of France (1667–1672)’s father?

Answer: Palace of Versailles

Question: Who is the paternal grandmother of Przemysław Potocki?

Answer: Ludwika Lubomirska

Question: Who lived longer, Herbert Findeisen or Léonie Humbert-Vignot?

Answer: Léonie Humbert-Vignot

Question: Are Alison Skipper and Diane Gilliam Fisher from the same country?

Answer: Yes

Question: Are director of film Move (1970 Film) and director of film Méditerranée (1963 Film) from the same country?

Answer: No
\\
\midrule
MQ & Question: Who is the child of the director and star of Awwal Number?

Answer: Suneil Anand

Question: What county shares a border with the county where Black Hawk Township is located?

Answer: Dodge County

Question: Who is the sibling of the person credited with the reinvention and popularization of oil paints?

Answer: Hubert Van Eyck

Question: Who heads the Catholic Church, in the country that a harp is associated with, as a lion is associated with the country that Queen Margaret and her son traveled to?

Answer: Eamon Martin
\\
\bottomrule
\end{tabular}
\caption{Demonstrations used in evaluation for each dataset}
\label{tab:eval_demos}
\end{table*}

\end{document}